\documentclass[11pt,a4paper]{article}
\usepackage[hyperref]{acl2019}
\usepackage{times}
\usepackage{latexsym}

\usepackage{url}
\usepackage{units}
\usepackage{amstext}
\usepackage{amsmath}
\usepackage{amsfonts}
\usepackage{amssymb}
\usepackage{adjustbox}
\usepackage{booktabs}
\usepackage{todonotes}
\usepackage{covington}
\usepackage{multirow}
\usepackage{multicol}
\usepackage[captionskip=-0.3cm]{subfig}
\usepackage{hhline}
\usepackage{siunitx}
\usepackage{arydshln}
\usepackage{enumitem}
\setlist[itemize]{noitemsep}
\setlist[enumerate]{noitemsep}
\setlist[description]{noitemsep}
\usepackage{makecell}
\usepackage{caption}

\aclfinalcopy 
\def\aclpaperid{4} 

\title{Second-order Co-occurrence Sensitivity of\\ Skip-Gram with Negative Sampling}

\author{Dominik Schlechtweg, Cennet Oguz, Sabine Schulte im Walde\\
    Institute for Natural Language Processing, University of Stuttgart \\
	{\tt \{schlecdk,oguzct,schulte\}@ims.uni-stuttgart.de}}

\date{}

\begin{document}
\maketitle
\begin{abstract}
We simulate first- and second-order context overlap and show that Skip-Gram with Negative Sampling is similar to Singular Value Decomposition in capturing second-order co-occurrence information, while Pointwise Mutual Information is agnostic to it. We support the results with an empirical study finding that the models react differently when provided with additional second-order information. Our findings reveal a basic property of Skip-Gram with Negative Sampling and point towards an explanation of its success on a variety of tasks.
\end{abstract}

\section{Introduction}
\label{sec:intro}

The idea of second-order co-occurrence vectors was introduced by \citet{Schutze1998} for word sense discrimination and has since then been extended and applied to a variety of tasks \cite{lemaire2006effects,islam2006second,SchulteImWalde:10,zhuang2018quantifying}. The basic idea is to represent a word $w$ not by a vector of the counts of context words it directly co-occurs with, but instead by a count vector of the context words of the context words, i.e., the second-order context words of $w$. These second-order vectors are supposed to be less sparse and more robust than first-order vectors \citep{Schutze1998}. Moreover, capturing second-order co-occurrence information can be seen as a way of generalization. To see this, cf. examples (\ref{ex:general1}) and (\ref{ex:general2}) inspired by \citet{hinjan93}.
\begin{examples}
\item \label{ex:general1} As far as the Soviet Communist \textbf{Party} and the Comintern were concerned \ldots 
\item \label{ex:general2} \ldots this is precisely the approach taken by the British \textbf{Government}.	
\end{examples}
The nouns \textit{Party} and \textit{Government} have similar meanings in these contexts, although they have little contextual overlap: A frequent topic in the British corpus used by \citeauthor{hinjan93} is the Communist Party of the Soviet Union, but governments are rarely qualified as communist in the corpus. Hence, there is little overlap in first-order context words of \textit{Party} and \textit{Government}. However, their context words \textit{Communist} and \textit{British} in turn have a greater overlap, because they are frequently used to qualify the same nouns from the political domain, as in \textit{Communist authorities} and \textit{British authorities}. Hence, although \textit{Party} and \textit{Government} may have no first-order context overlap, they do have second-order context overlap.
According to \citeauthor{hinjan93}, capturing this information corresponds to the generalization ``occurring with a political adjective.''

While most traditional count-based vector learning techniques such as raw count vectors or Point-wise Mutual Information (PPMI) do not capture second-order co-occurrence information, truncated Singular Value Decomposition (SVD) has been shown to do so. Regarding the more recently developed embeddings based on shallow neural networks, such as Skip-Gram with Negative Sampling (SGNS), it is presently unknown whether they capture higher-order co-occurrence information. So far, this question has been neglected as a research topic, although the answer is crucial to explain performance differences: \citet{Levy2015} show that SGNS performs similarly to SVD and differently from PPMI across semantic similarity data sets. If SGNS captures second-order co-occurrence information, this provides a possible explanation for the observed performance differences.

We examine this question in two experiments: (i) We create an artificial data set with target words displaying context overlap in different orders of co-occurrence and show that SGNS behaves similarly to SVD in capturing second-order co-occurrence information. The experiment supplies additional and striking evidence to prior work on SVD and introduces a method to further investigate questions more precisely than done before.
(ii) We transfer second-order context information to the first-order level in a small corpus and test the models' reaction on a standard evaluation data set when provided with the additional information. We find that SGNS and SVD, already capturing second-order information, do not benefit, whereas PPMI benefits.

\section{Related Work}
\label{sec:previous}

An early connection between second-order context information and generalization can be found in \citet{hinjan93}. The authors claim that SVD is able to generalize by using second-order context information as described above. Later work supports this claim and indicates that SVD even captures information from higher orders of co-occurrence \cite{landauer1998introduction,Kontostathis2002,NewoDipl05,Kontostathis2006}.

Since then, second-order co-occurrence information has mainly been exploited for traditional count-based vector learning techniques with different aims. \citet{Schutze1998} had used second-order vectors for word sense clustering. Various studies model synonymy or semantic similarity \citep{edmonds1997choosing,islam2006second,lemaire2006effects} indicating that second-order co-occurrence plays an important role for these tasks.

The only works we are aware of exploring second-order information for word embeddings are \citet{newman2017second} learning embeddings from nearest-neighbor graphs and \citet{zhuang2018quantifying} indicating that a specific type of word embeddings may benefit from second-order information. However, no study investigated the question whether SGNS or other word embeddings already capture higher-order co-occurrence information, which may make the integration of second-order information superfluous.

\section{Semantic Vector Spaces}
\label{sec:spaces}

We compare SGNS to two traditional count-based vector space learning techniques: PPMI and SVD, where the former does not capture second-order information while the latter does. All methods are based on the concept of semantic vector spaces: A semantic vector space constructed from a corpus $C$ with vocabulary $V$ is a matrix $M$, where each row vector represents a word $w$ in the vocabulary $V$ reflecting its co-occurrence statistics \citep{turney2010frequency}.

\paragraph{Positive Pointwise Mutual Information (PPMI).} For PPMI representations, we first construct a high-dimensional and sparse co-occurrence matrix $M$. The value of each matrix cell $M_{i,j}$ represents the number of co-occurrences of the word $w_i$ and the context $c_j$, $\#(w_i,c_j)$. Then, the co-occurrence counts in each matrix cell $M_{i,j}$ are weighted by the smoothed and shifted positive mutual information of target $w_i$ and context $c_j$ reflecting their degree of association. The values of the transformed matrix are

\vspace{-3mm}
\small
\begin{equation*}
{M}^{\textrm{PPMI}}_{i,j} = \max\left\lbrace\log\left(\frac{\#(w_i,c_j)\sum_c \#(c)^{\alpha}}{ \#(w_i)\#(c_j)^{\alpha}}\right)-\log(k),0\right\rbrace,
\end{equation*}
\normalsize
where $k > 1$ is a prior on the probability of observing an actual occurrence of $(w_i, c_j)$ and $0 < \alpha < 1$ is a smoothing parameter reducing PPMI's bias towards rare words \citep{Levy:2014,Levy2015}.
To our knowledge PPMI representations have never been claimed to capture higher-order co-occurrence information.

\vspace{+1mm}
\paragraph{Singular Value Decomposition (SVD).} Truncated Singular Value Decomposition is an algebraic algorithm finding the optimal rank $d$ factorization of matrix $M$ with respect to L2 loss \citep{Eckart1936}.\footnote{We use `SVD' to refer to the particular application of the algebraic method described.} It is used to obtain low-dimensional approximations of the PPMI representations by factorizing ${ M}^{\textrm{PPMI}}$ into the product of the three matrices ${ U}{ \Sigma}{ V}^\top$. We keep only the top $d$ elements of $\Sigma$ and obtain

\vspace{-2mm}
\begin{equation*}
{M}^{\textrm{SVD}} = { U_d}{ \Sigma^{p}_{d}},
\end{equation*}
where $p$ is an eigenvalue weighting parameter \cite{Levy2015}. Ignoring $V$ for ${M}^{\textrm{SVD}}$ reduces dimensionality while preserving the dot-products between rows.
While it is not clear whether SVD generalizes better than other models in general \citep{Gamallo2011}, its sensitivity to higher orders of co-occurrence has been shown empirically and mathematically \cite{Kontostathis2002,NewoDipl05,Kontostathis2006}.  \citet{Kontostathis2006} prove that the existence of a non-zero value in a truncated term-to-term co-occurrence matrix follows directly from the existence of a higher-order co-occurrence in the full matrix. They also show that there is an empirical correlation between the magnitude of the value and the number of higher-order co-occurrences found for the particular term pair.

\paragraph{Skip-Gram with Negative Sampling (SGNS).} SGNS differs from the above techniques in that it directly represents each word $w \in V$ and each context $c \in V$ as a $d$-dimensional vector by implicitly factorizing $M=WC^\top$ when solving

\vspace{-3mm}
\small
\begin{equation*}
\arg\max_\theta \sum_{(w,c)\in D} \log \sigma(v_c \cdot v_w) + \sum_{(w,c) \in D'} \log \sigma (-v_c \cdot v_w),
\end{equation*}
\normalsize
where $\sigma(x) = \frac{1}{1+e^{-x}}$, $D$ is the set of all observed word-context pairs and $D'$ is the set of randomly generated negative samples \citep{Mikolov13a,Mikolov13b,GoldbergL14}. The optimized parameters $\theta$ are $v_{c_i}=C_{i*}$ and $v_{w_i}=W_{i*}$ for $w,c\in V$, $i\in 1,...,d$. $D'$ is obtained by drawing $k$ contexts from the empirical unigram distribution $P(c) = \frac{\#(c)}{|D|}$ for each observation of $(w,c)$, cf. \citet{Levy2015}. 
The final SGNS matrix is given by

\vspace{-2mm}
\begin{equation*}
{M}^{\textrm{SGNS}} = W .
\end{equation*}
\citet{Levy:2014} relate SGNS to SVD by showing that under specific assumptions their learning objectives have the same global optimum. However, it is unknown whether SGNS is also similar to SVD in capturing higher-order co-occurrence information. The model architecture suggests that this is possible: consider the two context vectors $\vec{c_1},\vec{c_2}$ in $C$ of two words having large context overlap (e.g. the vectors for \textit{Communist} and \textit{British}). $\vec{c_1},\vec{c_2}$ will be similar, because the dot product with the same target vectors in $W$ will be maximized (as $\vec{c_1},\vec{c_2}$ frequently occur as contexts of the same target words). If $\vec{c_1},\vec{c_2}$ are then in turn each used to maximize the dot product with two different new target vectors (e.g. the vectors for \textit{Party} and \textit{Government}), these also tend to be similar.

\paragraph{Model Training.} For both experiments we use the implementation of \citet{Levy2015}, allowing us to train all models on extracted word-context pairs instead of the corpus directly. We follow previous work in setting the hyper-parameters \citep{Levy2015}. For PPMI we set $\alpha = .75$ and $k=5$. We set the number of dimensions $d$ for SVD and SGNS to 300. SGNS is trained with 5 negative samples, 5 epochs and without subsampling. For SVD we set $p=0$.

\paragraph{Similarity Measure.} For all methods we measure similarity between word vectors with Cosine Distance (CD), where low CD means high similarity. CD is based on cosine similarity, $cos(\vec{x},\vec{y})$, which measures the cosine of the angle between two non-zero vectors $\vec{x},\vec{y}$ with equal magnitudes \cite{salton1986introduction}. CD is then defined as
\begin{equation*}
CD(\vec{x},\vec{y})=1-cos(\vec{x},\vec{y}).
\end{equation*}

\vspace{+3mm}
\section{Experiment 1: Simulating second-order context overlap}

In order to see whether SGNS captures second-order co-occurrence information, we artificially simulate context overlap for first- and second-order co-occurrence separately. This allows us to simulate clear cases of overlap controlling for confounding factors which are present in empirical data. We generate target-context pairs in such a way that specific target words have either context word overlap in first-order co-occurrence, or by contrast in second order. We compare the behavior of PPMI, SVD and SGNS on three groups of such target words (see Table \ref{tab:simulpairs}): 

\begin{description}
    \item[first-order overlap (\textsc{1st}):] Target words $T$ occurring with the \emph{same} context words $C1$ in the first order, while in the second order all context words from $C1$ have distinct context words $C2$.
    \item[2nd-order overlap (\textsc{2nd}):] Target words $T$ occurring with \emph{distinct} context words $C1$ in the first order, while all context words from $C1$ have the same context words $C2$.
    \item[no overlap (\textsc{none}):] Target words $T$ occurring with distinct context words $C1$ in the first order and also all context words in $C1$ have distinct context words $C2$.
\end{description}

\noindent As an example, consider the column \textsc{2nd} in Table \ref{tab:simulpairs}. The target words $T$ are $a$ and $b$. Each has distinct context words: $a$ occurs only with $c,d \in C1$, while b occurs only with $e,f \in C1$. However, the first-order context words $c,d,e,f \in C1$ do have context overlap: $c,d,e,f$ occur all with $u,v \in C2$, i.e., they have the same second-order context words.

For each group we generate 10 target words ($T$). Per target word, each of $C1$, $C2$ is constructed by first generating 1000 context words $C$, assigning a sampling probability from a lognormal distribution to each context word in $C$ and then sampling 1000 times from $C$.\footnote{By sampling from a lognormal distribution we aim to approximate the empirical frequency distribution of context words. Context words receive probabilities by randomly sampling 1000 values from $f(x) = \frac{1}{x \sqrt{2\pi}}\exp\left(-\frac{\log(x)^2}{2}\right)$ and normalizing them to a probability distribution.} For the \textsc{1st}-group, the set of context words $C$ will be shared across target words, meaning that they have a context word overlap. For the target words in the \textsc{2nd}-group this will not be the case, but instead their first-order context words ($C1$) will have context overlap (see Table \ref{tab:simulpairs}). In this way, we simulate context overlap in first vs. second order. For the target words in the \textsc{none}-group, $C$ will instead be completely disjunct in both orders. Because co-occurrence is symmetric (if $a$ occurs with $c$, $c$ also occurs with $a$), for each pair ($a$,$c$) generated by the above-described process, we also add the reverse pair ($c$,$a$). To make sure that the pairs from the different groups (\textsc{1st}, \textsc{2nd}, \textsc{none}) do not interfere with each other, each string generated for a group is unique to the group. Finally, we mix and randomly shuffle the pairs from all groups. In this way, we generate 10 x 1000 x 1000 x 2 target-context pairs for each group: 10 target words occurring with 1000 context words in $C1$ where each in turn occurs with 1000 context words in $C2$, plus each of these pairs reversed.\footnote{Find the code generating the artificial pairs under: \url{https://github.com/Garrafao/SecondOrder}.}

Our main hypothesis is that SGNS and SVD will predict target words from the \textsc{2nd}-group to be more similar on average than target words from the \textsc{none}-group (although both groups have no first-order context overlap), while PPMI will predict similar averages for both groups.
 
\begin{table}[t]
	\center
	\small
	\begin{adjustbox}{width=0.65\linewidth}
	\begin{tabular}{ c |  c c c }
		\hline
	    \textbf{order}& \textbf{\textsc{1st}} & \textbf{\textsc{2nd}} & \textbf{\textsc{none}} \\
		\hline
		$\mathbf{C1}$ & \makecell{a \textbf{c}\\ a \textbf{d}\\b \textbf{c}\\ b \textbf{d}} & \makecell{a c\\ a d\\b e\\ b f} & \makecell{a c\\ a d\\b e\\ b f} \\
	    & & & \\
		$\mathbf{C2}$ & \makecell{c u\\ c v\\d w\\ d x} & \makecell{c \textbf{u}\\ c \textbf{v}\\d \textbf{u}\\ d \textbf{v}} & \makecell{c u\\ c v\\d w\\ d x} \\
		\hline
	\end{tabular}
	\end{adjustbox}

	\caption{Artificial co-occurrence pairs with context overlap in different orders of co-occurrence (\textsc{1st}, \textsc{2nd} and \textsc{none}). $C1$ and $C2$ give co-occurrence in first and second order respectively. For each pair (a,c) shown above we also add the reverse pair (c,a).} 
	\label{tab:simulpairs}
\end{table}

\begin{figure}
\includegraphics[width=1.0\linewidth]{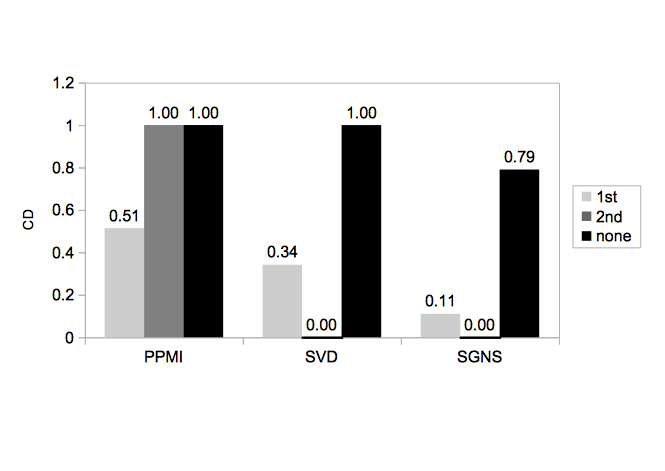}
\caption{Results of simulation experiment. Values give average cosine distances across target words with different levels of context overlap. Pair-wise differences between group means (except PPMI \textsc{2nd} vs. \textsc{none}) are statistically significant according to a two-sample bootstrap test ($p<0.001$, adjusted through Bonferroni correction for 9 tests, two-tailed).}
\label{fig:simulresults}
\end{figure}

\paragraph{Results.} Figure \ref{fig:simulresults} shows the average cosine distance between the target words in each of the three target word groups with context overlap in different orders (\textsc{1st}, \textsc{2nd} and \textsc{none}). As expected, PPMI predicts the target words without contextual overlap in any order (\textsc{none}) to be orthogonal to each other ($1.0$). Further, PPMI is sensitive to first-order overlap, but not at all to second-order overlap ($0.51$ vs. $1.0$). SVD also predicts orthogonality for the \textsc{none}-group ($1.0$) and shows sensitivity to first-order overlap ($0.34$), but is extremely sensitive to second-order overlap: it predicts the target words in \textsc{2nd} to be perfectly similar to each other ($0.0$), notwithstanding the fact that they have no first-order context word overlap. 
SGNS shows a similar behavior, although its vectors are distributed more densely: target words in \textsc{none} are predicted to be least similar ($0.79$), while target words in \textsc{1st} are more similar ($0.11$) and in \textsc{2nd} they are predicted to be completely similar ($0.0$).

We further hypothesize that the fact that for SGNS and SVD the average cosine distance in \textsc{1st} is higher than in \textsc{2nd} is related to our choice to make the context words $C1$ of the target words in \textsc{1st} dissimilar to each other by assigning completely distinct context words $C2$ (see Table \ref{tab:simulpairs}). We test this hypothesis by creating a second artificial set of target-context pairs completely parallel to the above-described set with the only difference that \textsc{1st} has additional context overlap in $C2$. On these targets words with overlap in both orders we find that PPMI makes similar predictions ($0.56$) as before, while for SVD and SGNS predictions drop to $0.0$, confirming our hypothesis.

\captionsetup{skip=-15pt}
\captionsetup{skip=0pt}

\paragraph{Discussion.} SGNS and SVD capture second-order co-occurrence information. Notably, they are more sensitive to the similarity of context words than to the words themselves (\textsc{2nd} vs. \textsc{1st}), which means that they abstract over mere co-occurrence with specific context words and take into account the co-occurrence structure of these words in the second order (and potentially higher orders). PPMI does not have this property and only measures context overlap in first order.

\section{Experiment 2: Propagating second-order co-occurrence information}

We now propagate second-order information to the first-order level by extracting second-order word-context pairs and adding them to the first-order pairs. We hypothesize that the additional second-order information will impact PPMI representations positively and stronger than SVD and SGNS, because we saw that the latter already capture second-order information. We reckon that the additional information is beneficial for PPMI in two ways: (i) it helps to generalize as described in (\ref{ex:general1}) and (\ref{ex:general2}), and (ii) it overcomes data sparsity for low-frequency words. Note that these two aspects are often highly related: with only a limited amount of data available it is more likely that similar words do not have exactly the same, but still similar context words. Generalization then helps to overcome sparsity.

\paragraph{Corpus.} We use ukWaC \cite{baroni2009wacky}, a ${>1}$B token web-crawled corpus of English. The corpus provides part-of-speech tagging and lemmatization. 
We keep only the lemmas which are tagged as nouns, adjectives or verbs. In order to assure we have low-frequency words in the test data, we create a small corpus by randomly choosing $1$M sentences and shuffling them. The final corpus contains roughly $10$M tokens. Under these sparse conditions we expect to observe strong effects on model performance highlighting the differences between the models.

\subsection{Pair Extraction} 

We first extract first-order word-context pairs by iterating over all sentences and extracting one word context pair ($w$,$c$) for each token $w$ and each context word $c$ surrounding $w$ in a symmetric window of size 5 (\textsc{base} pairs). Then, we extract additional second-order pairs in the following way: For each word type $t$ in the corpus, we build a second-order vector $\vec{v}$ by summing over all of $t$'s first-order context token count vectors \cite{Schutze1998}. Then we randomly sample $n$ second-order context tokens from $\vec{v}$ (with replacement) where each context type $c_i$ has a sampling probability of $\frac{\vec{v}_i}{\sum_{j=1} \vec{v}_j}$ and $\vec{v}_i$ is $\vec{v}$'s $i$th entry. We then exclude all sampled context tokens $c = t$.\footnote{Find the code under: \url{https://github.com/Garrafao/SecondOrder}.}

We extract second-order pairs only for words below a specific co-occurrence frequency threshold $f$ to test the impact on sparse words separately. We experiment with $f\in\{$2k,20k,200k$\}$. We set $n$ globally to $200\%$ of $t$'s co-occurrence frequency to add a substantial amount of information. For each of the second-order pairs we add the reverse pair ($c$,$w$). Finally, the respective second-order pairs are combined with the base pairs and randomly shuffled. In this way, we generate roughly 22/99/218M (2/20/200k) second-order pairs from 57M base pairs. Then we train each model on each of the combined pair files separately.

\subsection{Results}

\captionsetup{skip=-15pt}
\begin{figure}[t]
   \includegraphics[width=\linewidth]{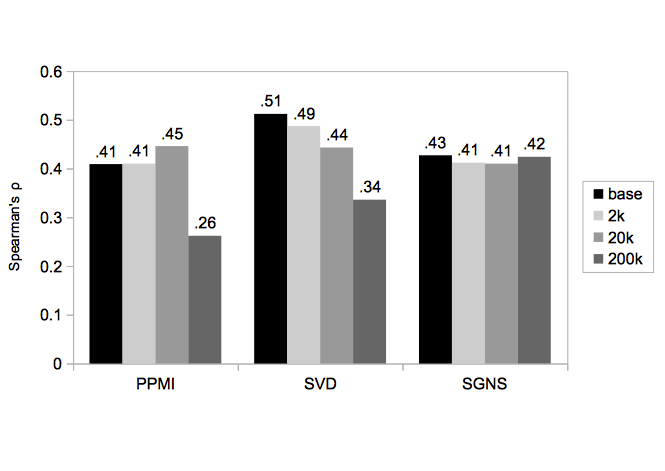}
    \caption{Results of experiment 2. Values give correlation (Spearman's $\rho$) of model predictions with human similarity judgments.}
    \label{fig:pairsresults}
\end{figure}

We evaluate the obtained vector spaces on WordSim353 \citep{finkelstein2002placing}, a standard human similarity judgment data set, by measuring the Spearman correlation of the cosine similarity for target word pairs with human judgments. The results are shown in Figure \ref{fig:pairsresults}. As we can see, the models show different reactions to the additional second-order information: PPMI is the only model benefiting (in one case), while SVD and SGNS never benefit from the additional information and are always impacted negatively. The strongest negative impact can be observed for SVD (-0.17). PPMI shows a clear improvement with rather low-frequency words (20k). Adding second-order information for high-frequency words (200k) has a strong negative impact for PPMI and SVD.

\paragraph{Discussion.} The different reactions of PPMI vs. SVD and SGNS partly confirm our hypothesis which was based on the findings in experiment 1: only PPMI benefits from additional second-order information. However, we did not expect the observed negative impacts, especially the strong performance drop for SVD. Moreover, it is notable that SVD on the base pairs shows a much higher performance ($0.51$) than SGNS and PPMI ($0.43$, $0.41$), which is not the case in less sparse conditions \cite{Levy2015}. This indicates that SVD makes much better use of the available information and overcomes data sparsity in this way. It remains for future research to determine how much the exploitation of higher-order co-occurrence information contributes to this clear performance advantage.

\vspace{+4mm}
\section{Conclusion}
\label{sec:conclusion}

We showed that SGNS captures second-order co-occurrence information, a property it shares with SVD and distinguishes it from PPMI. We further tested the reaction of the models when provided with additional second-order information, expecting that only PPMI would benefit. We find this confirmed, but also observe unexpectedly strong negative impacts on SVD by the supposedly redundant information. In general, SVD turns out to have strong performance advantages over PPMI and SGNS in the sparse experimental conditions we created.

Our findings are relevant for a variety of algorithms relying on the SGNS architecture \citep[i.a.][]{Grover2016,bamler17}. Future work will look into the relationship between the second-order sensitivity of SVD and SGNS and their high performances across tasks. In addition, we aim to use the introduced method of generating artificial context overlap to see which higher orders of co-occurrence SVD, SGNS and other embedding types \cite{pennington-etal-2014-glove,peters-etal-2018-deep,athi_multift_2018} capture. Because the aim of the study was only to test the second-order sensitivity of different models, we did not focus on finding the best way to provide this information. Given the results for PPMI, however, developing a smoother way to provide second-order information to models seems to be a promising starting point for further research.

\vspace{+5mm}
\section*{Acknowledgments}

The first author was supported by the Konrad Adenauer Foundation and the CRETA center funded by the German Ministry for Education and Research (BMBF) during the conduct of this study. We thank Agnieszka Falenska, Sascha Schlechtweg and the anonymous reviewers for their valuable comments.

\vspace{+3mm}
\bibliography{190316-bbnlp-sos}
\bibliographystyle{acl_natbib}

\end{document}